# Enhancement of the Color Image Compression Using a New Algorithm based on Discrete Hermite Wavelet Transform

*Hassan Mohamed Muhi-Aldeen*
*Department of Computer Engineering*
*Aliraqia University*
*22Sabaabkar, Adamia, Baghdad, Iraq*
*muhialdeen.hassan@aliraqia.edu.iq*
*https://orcid.org/0000-0002-9177-8621*

*Asma A. Abdulrahman*
*Department of Applied Sciences*
*University of Technology*
*52Alsena street, Baghdad, Iraq*
*Asma. A. Abdulrahman@uotechnology.edu.iq*
*http://orcid.org/0000-0001-6034-8859*

*Jabbar Abed Eleiwy*
*Department of Applied Sciences*
*University of Technology*
*52Alsena street Baghdad, Iraq*
*jabar.a.eleiwy@uotechnology.edu.iq*
*https://orcid.org/0000-0002-7442-1719*

*Fouad S. Tahir*
*Department of Applied Sciences*
*University of Technology, Iraq*
*52Alsena street, Baghdad, Iraq*
*fouad.s.taher@uotechnology.edu.iq*
*http://orcid.org/0000-0002-3121-201X*

*Yurii Khlaponin*
*Corresponding author*
*Department of Cybersecurity and Computer Engineering*
*Kyiv National University of Construction and Architecture*
*31 Povitroflotskyi ave, Kyiv, Ukraine, 03037*
*y.khlaponin@knuba.edu.ua*

**Abstract:** The Internet has turned the entire world into a small village; this is because it has made it possible to share millions of images and videos. However, sending and receiving a huge amount of data is considered to be a main challenge. To address this issue, a new algorithm is required to reduce image bits and represent the data in a compressed form. Nevertheless, image compression is an important application for transferring large files and images. This requires appropriate and efficient transfers in this field to achieve the task and reach the best results. In this work, we propose a new algorithm based on discrete Hermite wavelets transformation (DHWT) that shows the efficiency and quality of the color images. By compressing the color image, this method analyzes it and divides it into approximate coefficients and detail coefficients after adding the wavelets into MATLAB. With Multi-Resolution Analyses (MRA), the appropriate filter is derived, and the mathematical aspects prove to be validated by testing a new filter and performing its operation. After the decomposition of the rows and upon the process of the reconstruction, taking the inverse of the filter and dealing with the columns of the matrix, the original matrix is improved by measuring the parameters of the image to achieve the best quality of the resulting image, such as the peak signal-to-noise ratio (PSNR), compression ratio (CR), bits per pixel (BPP), and mean square error (MSE).

## 1. Introduction

By developing a wavelet transformation, it was possible to simplify the process of image analysis and compression. This wavelet transformation proved particularly useful for DWT, which gained more popularity than its forerunners. Similar to DCT, this is because DWT uses less energy due to the continuity property it possesses, but DCT produces compressed images that are superior to DWT's [1–5]. Quality is used to gauge how well an image was compressed. Even if there is a loss, it will not be visible if there is no loss in the resulting image information [6–10]. When the losing compression is contrasted with the non-losing compression, the former has high compressed ratios, while the second has a small, imperceptible difference. Pixels in an image are converted into frequency coefficients by discrete Fourier transform (DFT) and discrete cosine transform (DCT), which results in the loss of image data. Wavelets are used in the analysis of the color image to produce highly accurate images, or smooth images are used instead [11–15]. Polynomials will soon be used to obtain approximations of results. The image information was maintained while achieving high compression rates [16–20]. The image is transformed into a two-dimensional array and a set of data [21–25] from the mathematical perspective of the image data shrinkage process. Due to the advancement of computers and the requirement to store data for a single person's color images on a web page via the Internet, a catalog with hundreds of images must logically be compressed in order to store the data that needs enough space for storage. JPEG is the common file type used to store compressed images. In order to achieve some compression during the quantification process, DCT divides images into segments with various frequencies. [26]. Duplicating data that needs high frequencies can be difficult for wireless sensor networks. These issues are resolved by image compression technology. [27]. To get a rough idea of the dimensions of the data, use arrays (NMF). We obtained the compression systems SVD, NMF, and PNMF. send picture images through it. Image compression in the medical field aims to shrink the file size without sacrificing the original image's quality. There are digital pictures. DICOM is the industry standard for a medical imaging file to store patient information. The data is compressed using Hoffman lossless compression rather than DICOM [28]. Thanks to the wavelet function, which offers a high compression ratio without sacrificing the original data, the color image was compressed using wavelets and the EZW algorithm to produce better compression results. [29], [30-31]. In this study, a color image was compressed using the discrete Hermite wavelet transform (DHWT). After the wavelets were added to the MATLAB program, it was analyzed and divided into approximate coefficients and detailed coefficients. With Multi-Resolution Analyses (MRA), the proper filter was derived, the mathematical components were verified, and the filter was tested by using a matrix and running an operation. The original matrix was obtained through decomposition with rows, reconstruction with the inverse of the filter, and handling of the matrix's columns. A new algorithm was created to incorporate the new filter into the MATLAB program, making the wavelet ready for use alongside the other common wavelet wavelets in the program, including Symlet2, Coiflet 2, and Daubecheis2. Peak signal-to-noise ratio (PSNR), compression ratio (CR), bits per pixel (BPP), and mean square error were measured after the image was truly compressed (TC) using the algorithms (MSE). The original color image in Figure 1 is analyzed using DHWT, and the outcomes are contrasted with a variety of basic wavelets. The effectiveness and caliber of the suggested new wavelets are shown in Tables (2) and (3).

## 2. Discrete Chebyshev Wavelet Transform (DCHWT)

In this section, the input image is analyzed using Discrete Chebyshev Wavelet Transform (DCHWT) for the purpose of frequency extraction and as shown in Figure (1) with an effective technique in order to preserve the details of the image being processed and use this method in many areas such as edge detection, image compression and color coding.

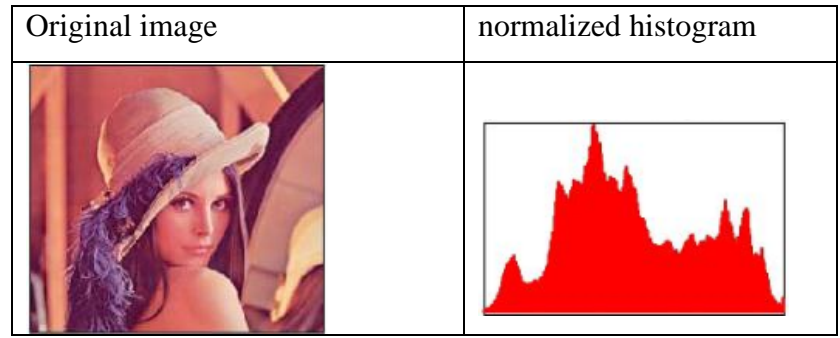

**FIGURE 1: shows the original color image that is analyzed using DHWT**

## 2. Preliminaries

This section includes all the basic and important equations that will contribute to the completion of this work

### Hermite Wavelets Transform [11-14]

An important equation which appears in problems of physics is called Hermite differential equation; it is given by

$$y'' - 2xy' + 2ny = 0 \qquad (1)$$

where n=0,1,2,3…

Equation (1) has polynomial solutions called Hermite polynomials given by Rodrigue's formula

$$H_n(x) = (-1)^n e^{x^2} \frac{d^n}{dx^n}\left(e^{-x^2}\right) \qquad (2)$$

- The first few Hermite polynomials are
  $H_0 = 1, H_1 = 2x, H_2 = 4x^2 - 2, H_3 = 8x^3 - 12x$
- The generating function for Hermite polynomials is given by

$e^{2tx-t^2} = \sum_{n=0}^{\infty} \frac{H_n}{n!} t^n$

This result is useful in obtaining many properties of $H_n(x)$. The Hermite polynomials satisfy the recurrence formulas

$$H_{n+1}(x) = 2xH_n(x) - 2nH_{n-1}(x) \tag{3}$$

$H'_n(x) = 2nH_{n-1}(x)$

Starting with $H_0 = 1, H_1 = 2x$.

Orthgonality of Hermite polynomials

$$\int_{-\infty}^{\infty} e^{-x^2} H_m(x) H_n(x) dx = \begin{cases} 0 & m \neq n \\ 2^n n! \sqrt{\pi} & m = n \end{cases} \tag{4}$$

So that the Hermite polynomials are mutually orthogonal with respect to the weight function or density function $e^{-x^2}$ and if m=n we can normalize the Hermite polynomial so as to obtain an orthonormals set.

A lot of research included how to build the wavelet, which consists of the mother function, where the movement depends on two important coefficients a and b, through which the extension and the first translation is responsible for the extension and the second of the translation where the process is continuing on this case:

$$h_{a,b}(x) = |a|^{\frac{-1}{2}} h\left[\frac{x-b}{a}\right] \quad a,b \in R, \ a \neq 0 \tag{5}$$

Let dilation by parameter $a = 2^{-k}$, translation by parameter $b = (a(2n-1))$ and transform $x = a(2a^{-1}t)$, by substitute parameters a, b and transform x in equation (5), then will be get equation (6). DHWT $h_{n,m}(t) = h(k,n,m,t)$ include four parameters, $n = 1,2,...,2^k$ k is assumed any positive integer, m is the degree of the Hermite polynomials and t independent variable in $[0,1]$.

$$h_{n,m}(t) = \begin{cases} 2^{\frac{k}{2}} H_m^*(2^{k+1}t - 2n + 1) & t \in \left[\frac{n-1}{2^k}, \frac{n}{2^k}\right] \\ 0 & otherwise \end{cases} \tag{6}$$

where $$H_m^* = \frac{1}{2^m m! \sqrt{\pi}} H_m \tag{7}$$

$m = 0,1,2,..., M-1 \qquad n = 0,1,2,...,2^k$

A function f(t) defined over [0,1) may be expanded as a function f(t) defined over [0,1) may be expanded as:

$$f(t) = \sum_{n=1}^{\infty} \sum_{m=0}^{\infty} c_{n,m} h_{n,m}(t) \tag{8}$$

where $\quad C_{n,m} = (f(t), h_{n,m}(t)) = \int_{0}^{1} w_n(t) h_{n,m}(t) f(t) dt$

in which $\langle .,. \rangle$ denoted the inner product in $L^2_{w_n}[0,1]$.

If the infinite series in equation(8) is truncated, then it can be written as:

$$f(t) \cong f_{2^k, M-1} \sum_{n=1}^{2^k} \sum_{m=0}^{M-1} C_{n,m} h_{n,m}(t) = C^T h(t) \tag{9}$$

where F and $h(t)$ are $2^k M \times 1$ matrices given by:

$$h(t) = \left[ h_{1,0}(t), h_{1,1}(t), ..., h_{1,M-1}(t), h_{2,0}(t), ..., h_{2,M-1}(t), ..., h_{2^k,0}(t), ..., h_{2^k,M-1}(t) \right]^T$$

$$C = \left[ C_{1,0}, C_{1,1}, ..., C_{1,M-1}, C_{2,0}, ..., C_{2,M-1}, ..., C_{2^k,0}, ..., C_{2^k,M-1} \right]^T \tag{10}$$

## 3. Discrete Hermite Wavelets Transformation (DHWT)

In this section, the detailed mathematical aspects of the proposed Discrete Hermite Wavelet Transform (DHWT) are addressed, in order to study its effect in designing a digital watermarking image algorithm later in this work.

### 3.1 Mathematical Derivation of Discrete Hermite Wavelet Transform (DHWT)

In this section the mathematical derivation of the DHWT is introduced. The time $t$ and frequency $\zeta$ respectively is in the space $L^2(R)$ and Hermite transform is $H_n$ and $\overline{H}_n$ is the invers of Hermite transform. The following equations represent analysis and synthesis of Hermite Wavelet Transform respectively

$$A(\zeta) = (H_n f)(\zeta) = \frac{1}{(2^m m! \sqrt{\pi})^2} \int_R f(t) H_n(t) e^{-(\zeta t)^2} dt \quad \zeta \in R$$

$$f(t) = (\overline{H}_n A)(t) = \int_R A(\zeta) e^{-(\zeta t)^2} d\zeta \quad t \in R$$

The series of $f$ will be $A(\zeta)$, then the energy series is $|A(\zeta)|$ we say the signal is linear and continuous then $H_n$ is continuous and linearly independent application of $L^2(R)$ then $\overline{H}_n \Rightarrow f = \overline{H}_n H_n f = H_n \overline{H}_n f$ in the norm space the inner product of same properties

$\langle f, p \rangle_{L^2} = \langle H_n f, H_n p \rangle = \langle \overline{H}_n f, \overline{H}_n p \rangle$ and the norm form $\|f\|_{L^2} = \|H_n f\|_{L^2} = \|\overline{H}_n f\|_{L^2}$

The atoms of this transform are $e^{-(\zeta t)^2}$, It seems that in this analysis will get some negatives of Hermite transform. If $f$ is not taken continuously, the temporal parts of this funtion will fade, making it impossible to test using the coefficients, Hermie transform will lose their cause, and The principle of doubt for Heisenberg see[].

### 3.2 Analysis of the Discrete Hermite Wavelet Transform (DHWT):

The Hermite wavelet is a regular and well defined function because $h \in L^1 \cap L^2$, and $\int_R \frac{|h(\zeta)|^2}{\zeta} d\zeta \in [-\infty, \infty]$ is the frequency domain, where $h$ is the signal of the wavelet $h_{n,m}$. The Hermite wavelet condition is in the period $[0,1]$, m is the vanishing moments by

$\int_R t^s h(t) dt = 0$, $s = 0,1,2,...,m$ as long as $h \in L^1 \cap L^2$, $t \in L^1$, $\int_R h(t) dt = 0$, is the admissibility condition.

The temporal factor plays a major role in narrowing and extending the wavelet due to fading. This means that the wavelet can be divided by the number of times it disappears.

The Hermite wavelet transform, with $f \in R$, with time $t$ can be defined the coefficients

$$A_{a,b} = \int_R f(t) h_{a,b}(t) dt \quad a \in R^+, b \in R$$

Function $f$ analysis is equal to the calculation of coefficients, and the importance of this process in measuring the fluctuations of the function by a scale a, Large wave transaction values are with the scale b, the scaling function denoted by $\omega$ with wavelet $h$ by used scaling function in orthogonal with wavelet function $h$ can be get the two basic wavelets are called Hermite wavelets

$$\begin{cases} h_{n,m}(t) = 2^{\frac{k}{2}} h(2^{k+1} t - n + 1) \\ \omega_{n,m}(t) = 2^{\frac{k}{2}} \omega(2^{k+1} t - n + 1) \\ for\ (n,m) \in Z^2 \end{cases}$$

The wavelet coefficients of a signal S are given by

$$\delta_{n,m} = \int_R S(t) h_{n,m}(t) dt$$

$$\eta_{n,m} = \int_R S(t) \omega_{n,m}(t) dt$$

The reconstructed the signal

$$S(t) = \sum_{n \in Z} \sum_{m \in Z} h_{n,m}(t) \delta_{n,m}$$

The concept of Multi Resolution Analysis (MRA) Summarize in the orthonormal family of basis wavelet $\{h_{n,m}\}$ $(n,m) \in Z^2$ in $L^2(R)$ in $\{V_n\}_{n \in Z}$ is a finite sequence $... \subset V_2 \subset V_1 \subset V_0 \subset V_{-1} \subset V_{-2} \subset ...$ in $L^2(R)$ $V_0$ is center of these spaces $n < 0$, the scaling function $\omega_{n,m}$ generates of $V_0$ with $f(t) \in V_n \Leftrightarrow f(2t) \in V_{n-1}$, be defined

$$V_0 = \left\{ f \in L^2(R) \middle| f(t) = \sum_{n \in Z} A_{n,m} \omega(2^{k+1} t - 2n + 1) \in l^2(Z) \right\}$$

The MRA of DHWT, the subspace $\{V_n\}_{n \in Z}$ of MRA is approximation spaces with scaling function $\omega_{n,m}$, with detail spaces denoted by $\{W_n\}_{n \in Z}$, the space $W_n \perp V_n$ in $V_{n-1} = V_n \oplus W_n$, $n \in Z$, $\{\omega_{n,m}\}$ $(n,m) \in Z^2$ is generated $\{V_n\}_{n \in Z}$ and $\{h_{n,m}\}$ $(n,m) \in Z^2$ is generated $\{W_n\}_{n \in Z}$, the coefficient $\delta_{n,m}$ is the orthogonal projection of S on $\{W_n\}_{n \in Z}$.

In the coming sections will knew how to use Discrete Hermite Wavelets Transformation DHWT to analyze and reconstruct the signal.

### 3.3 The Coefficients of the Discrete Hermite Wavelet Transform (DHWT)

The approximation coefficients of DHWT is
$$\eta_{n,m} = \int_R S(t) \omega_{n,m}(t) dt \text{ where } A_n(t) = \sum_{m \in Z} \eta_{n,m} \omega_{n,m}(t)$$

The detail coefficients of DHWT is
$$\delta_{n,m} = \int_R S(t) h_{n,m}(t) dt \text{ where } D_n(t) = \sum_{m \in Z} \delta_{n,m} h_{n,m}(t)$$

$\eta_{n,m}, \delta_{n,m}$ are coefficients of $S$, in level n with the coefficients $\{\eta_{0,m}\}_{m \in Z} \in V_0$ and $A_n \subset V_n$, $D_n \subset W_n$ of $S$.

From the transaction can be found decision tree started from the level, $n = 0$, $A_0 = S$, of The following steps will be obtained

Level 1: $D_1 = A_0 - A_1 = S - A_1 \Rightarrow S = D_1 + A_1$

Level 2: $D_2 = A_1 - A_2 = S - D_1 - A_2 \Rightarrow S = D_2 + D_1 + A_2$

$$S = \sum_{n=0}^{2^k} \sum_{m=0}^{M-1} \delta_{n,m} h_{n,m}(t)$$

Generally at level N using the above equations the following equation can be obtained

$$S = A_N + \sum_{n \leq N} D_n \Rightarrow A_{N-1} = A_N + D_N \Rightarrow \{h_{n,m}\}_{n,m \in Z}, \text{ is orthogonal.}$$

3.4 The Discrete Hermite Wavelet Transform Packets (DHWTP):

In this section the role of orthogonal beams will be displayed by approximate coefficients and details, the following steps will demonstrate how to create DHWTP.

Step 1: scaling function $\omega \in V_0$ in MRA $\omega_{0,m} = \omega(t-m) \Rightarrow \exists s = \{s_m\}_{m \in Z} \in l^2$ such that

$$\frac{1}{2}\omega\left(\frac{1}{2}\right) = \sum_{m \in Z} s_m \omega(t-m) \in L^2$$

Wavelet function $h \in W_0 \Rightarrow h_{0,m} = h(t-m) \Rightarrow \exists r = \{r_m\}_{m \in Z} \in l^2$ such that

$$\frac{1}{2}h\left(\frac{1}{2}\right) = \sum_{m \in Z} r_m \omega(t-m) \in L^2$$

The orthonormal wavelet in the interval $[0,1)$, $\{h_{n,m}\}_{n,m \in Z} \in L^2(0,1) \subset L^2(R)$

Step 2: the two filters $y_m$, $u_m$ with length 2M related with wavelet and scaling function with orthogonally $h, \omega$ respectively.

Step 3: basis sequences $\{s_m\}_{m \in Z}$, $\{r_m\}_{m \in Z} \subset l^2$ norm equal 1.

Step 4: defined two functions $H_n$, $n \in Z$, $H_0 = \omega$, the scaling function

$$\begin{cases} H_{2m}(t) = \sqrt{\pi} u_m H_m(2t-n) \\ H_{2m+1}(t) = \sqrt{\pi} y_m H_m(2t-n) \end{cases}$$

The scaling function m=o $H_0 = \omega$, $H_1 = h$ is the Hermite wavelet function by use above equation we get to $u_0 = u_1 = \dfrac{1}{\sqrt{\pi}}$, $y_0 = -y_1 = \dfrac{1}{\sqrt{\pi}}$

Step 5: if $H_0$ is the scaling function and $H_1$ is the Hermite wavelet in $[0,1)$ $H_m(t)$, $t \in [0,0.5)$ obtained $H_m(2t)$, then obtained $H_{2m}$, and in the interval $[0.5,1)$, $H_m(2t-1)$ and $H_{2m+1}$.

From above obtained the atoms of HWT from the dilation and translation for the scaling function $\omega$ and Hermite wavelet $h$, then DHWTP in the following equation

$$(H_m)_{k,n}(t) = 2^{\frac{-k}{2}} H_m(2^k t - n), \; for\, m \in N, \, (k,n) \in Z^2, \, n \in [0, 2^k - 1]$$

The set function $h_{k,n} = (H_m)_{k,n}(t)$ the Discrete Hermite Wavelet Transform Packets (DHWTP), by wavelets of DHWTP $h_{k,n}$ and $\bar{h}_{n,k}$ $\forall k$, $\bar{h}_{k,0} = V_k$, and $\bar{h}_{k,1} = W_k$ with $V_0 = \bar{h}_{0,0} \Rightarrow h_{r,1} : r \geq 1$, are orthogonal base of $V_0$, $N \geq 0$, $(h_{N,0}, (h_{k,1}; k \in [1, N]))$, is orthogonal base of $V_0$, the following regular of DHWTP

1- For $k = 0$, will be get $h_{0,0}$
2- For $k = 1$, from $h_{0,0}$ will be get two functions $h_{1,0}$ and $h_{1,1}$
3- For $k = 2$ from the above steps will be get the following functions $h_{2,0}, h_{2,1}, h_{2,2}, h_{2,3}$

In general level k will be get from $h_{k,n}$ the following functions $h_{k+1,2n}$, $h_{k+1,2n+1}$

**TABLE 1. Shows the DHWT K=0,1,2** $t \in [0,1)$

$$\omega_{k,n}(t) = \begin{cases} \dfrac{1}{\pi} & 0 \leq t < 1 \\ 0 & otherwise \end{cases}$$

|  | K=0 | K=2 | K=2 |
|---|---|---|---|
| $h_{0,0}$ | $\dfrac{1}{\sqrt{\pi}}$ | 0 | 0 |
| $h_{1,0}$ | 0 | $\dfrac{1}{\sqrt{\pi}}$ | 0 |
| $h_{1,1}$ | 0 | $\frac{1}{\sqrt{\pi}}(2(2t-1))$ | 0 |
| $h_{2,0}$ | 0 | 0 | $\frac{1}{\sqrt{\pi}}$ |
| $h_{2,1}$ | 0 | 0 | $\frac{1}{\sqrt{\pi}}(2(4t-1))$ |
| $h_{2,2}$ | 0 | 0 | $\frac{1}{\sqrt{\pi}}(4(4t-2)^2 - 2)$ |
| $h_{2,3}$ | 0 | 0 | $\frac{1}{\sqrt{\pi}}(8(4t-3)^3 - 12(4t-3))$ |

DHWT is applied on a $4 \times 4$ sub-image for more illustration. The coefficients of DHWT obtained above are used where the filter extracted is

$(2 \times 2)$ F$=\frac{1}{\sqrt{\pi}}\begin{bmatrix}1 & -1\\1 & 1\end{bmatrix}$ in figure (2)

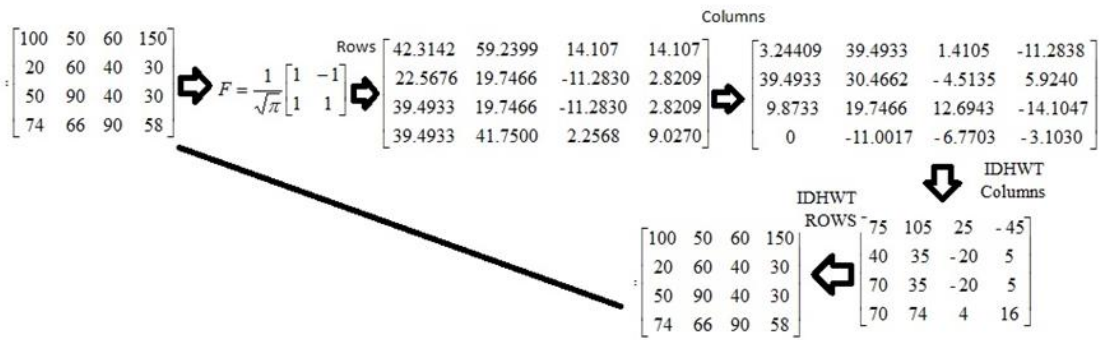

**FIGURE 2: Multi Resolution Analyses (MRA) with DHWT**

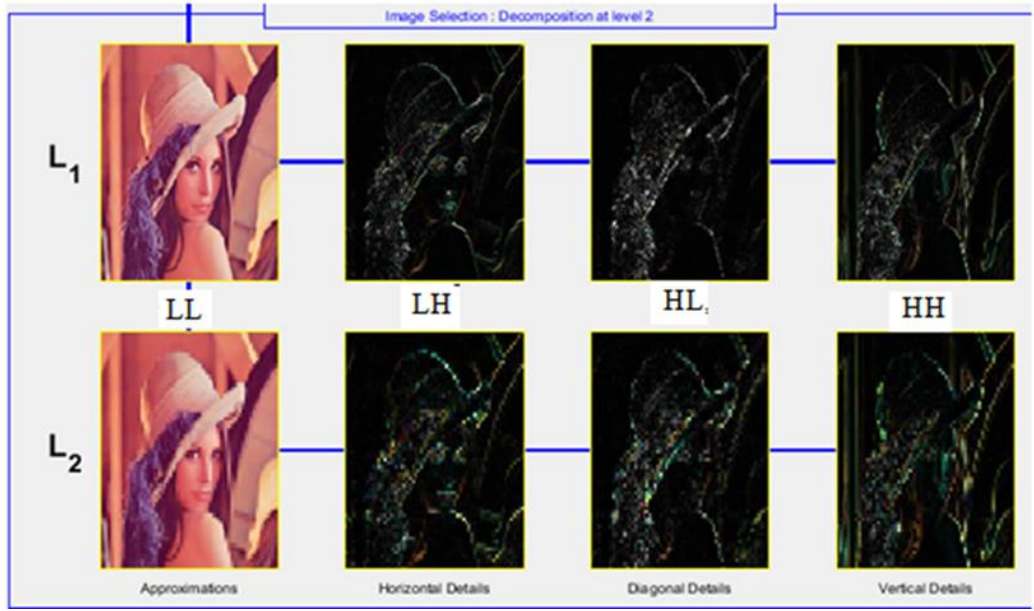

**FIGURE 3 DHWT Effect on Lena gray Image**

## 4. MATLAB

**Add DHWT in MATLAB**

In this section, an algorithm was developed to show how to add the proposed wave in the MATLAB, which requires this work before the start of the algorithm classification of any wavelet family mentioned in the first chapter, which enters the field Wavelet Toolbox (WT) using the "wavemngr" function in Matlab program.

**The Stages of Creating a Waveform Table contains DHWT**

**Algorithm(1) Add New DHWT in MATLAB:** Create an orthogonal wavelet family containing only one wavelet.

Step 1: Create a Wavelet Family with One Type 1 Wavelet Orthogonal wavelet With scale function

Step 2: Define the five parameters associated with the wavelet family. And use the "wavemngr" function

Step 3: Write the following steps in order to determine the wave of the first type is the orthogonal type using the following function

>> familyName = 'MyWAVE T1';

Step 4: In this step short name must be specified familyShortName = 'her';

familyWaveType = 1;

familyNums = '';

fileWaveName = ' her.mat';

Step 5: In this step need great filter associated with the wavelet and save it in the

MAT-file mywa.mat. This step must be done before using the new wavelet

family. This filter will be used by the "wfilters" function to build four filters

used by the Discrete Hermite Wavelet Transform IDHWT  and save it

\>\> save her her.

Step 6: In this step, the new wavelet family will be added to a pile of captive families.

Step 7: Finally, make sure to add the new wavelet through the instructions

## 5. Compression Image by Discrete Hermit Wavelet  Transform (DHWT)

In this section, the focus is on aspects related to the principles of compression image (CI) using the proposed theory, where a compressed image is obtained. The algorithm 3.5 illustrates the stages of the process implementation Figure3.8: shows the steps of the algorithm

**Algorithm (2): compressing color image by DHWT Algorithm by level threshold method**

Input: Color Image in size $(256 \times 256)$

Output: Compressed Image (CI)

Step1: load color image

Step2: The wavelet transform DHWT is It is inserted on the image.  The color map is

  smooth

Step3: The wavelet transform DHWT with image divided to two   Coefficients

   Approximate  Coefficients (AC) and Details Coefficients (DC)

Step4 : Discrete Hermit Wavelet Transform (DHWT), on the  image in the first

  level, the image is divided into four blocks each block $(64 \times 64)$ $LL_1$, $LH_1$,

   $HL_1$, $HH_1$ In the level2, the first quarter $LL_1$ is divided into four blocks each

block (16 × 16), $LL_2$, $LH_2$, $HL_2$, $HH_2$

Step5: For each level, the threshold is determined from the detail parameters, that is, to the mentioned factors, the fixed threshold is applied

Step6: The image is compressed at this stage for the purpose of storage or transportation after the completion of the process.

Step7: The reconstruct image by invers Discrete Hermit Wavelet Transform (IDHWT) Regenerate an image.( Re-creation of wavelets DHWT through the original approximate coefficients for level 2 on the other hand. The modified detail coefficients at level 1 and 2)

Step8: The multi-level wavelet of DHWT re-decomposes steps 3 and 6

Figure Four shows the threshold by a level method selected between -100 and 100 in the horizontal, diagonal, and vertical detail coefficients in a compressed image, which returned 99.96% energy and a number of zero for 29.11% using the proposed method, the Discrete Hermit Wavelet Transform.

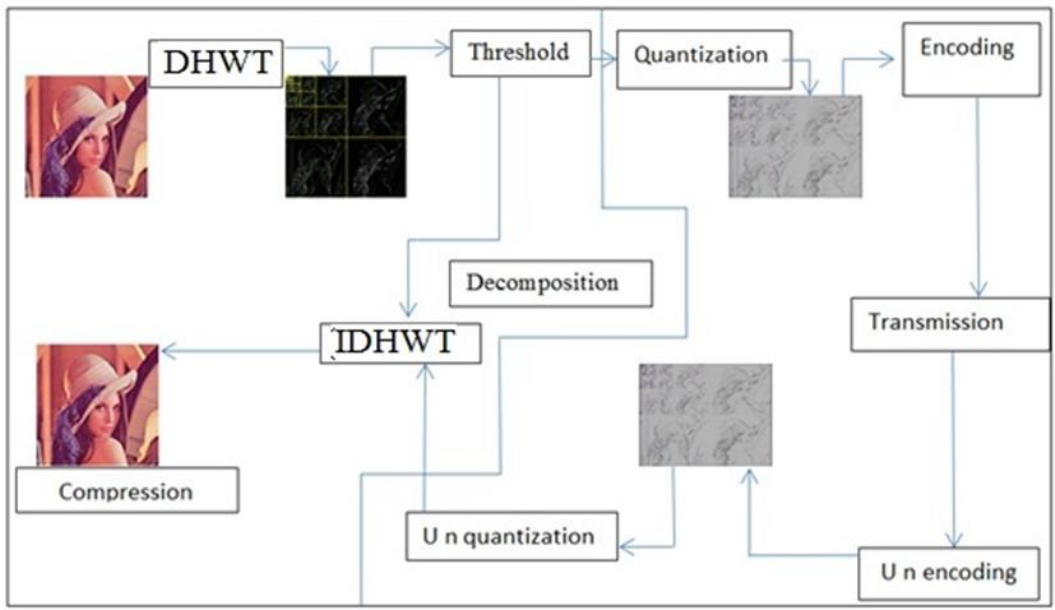

**FIGURE 4:  shows the stages of compression in algorithm**

The algorithm represents a type of compression that causes loss of information during two stages.

1. When adjusting some parameters in the threshold step

2. When deducting the value of certain coefficients in a quantification step

Once the threshold has been removed, the error has also been completely removed, and a role for DHWT with integer or rational coefficients has been revealed. The role of DHWT appears in the first step of compression when analyzing the image, and in the final step, when rebuilding the image using an inverse IDHWT. The tree structure of DHWT has a role in coding and focuses on the boundaries that connect DHWT, the threshold, and IDHWT.

## 7. The Results

Four samples were used to submit the proposed algorithms 1 and 2, and the intermittent DHWT decomposition was used to analyze these samples. On level 8 for decomposition, using the zero-coding tree shows the ability of the algorithm to evaluate MSE, PSNR, CR, and B.P.P. Tables show MSE the quality of the reconfigured image, with the average squared error equal to zero. For good results, the peak signal-to-noise ratio (PSNR) should be high. The compression capacity can be analyzed using important metrics for image quality.

To assess image compression, quality measures are used. Data transfer and image compression depend on compression ratio and PSNR.

For storing and viewing photos, it requires more memory, per pixel of the image. The picture quality is an increase in the number of bits to represent more color, and in order to view and store the image, more memory is required, which is the error of the difference from the original image. with the resulting image G serving as a decoder for u, v (the sum of all pixels indicating variance and the change in the rebuilding error). PSNR is given between two 8-bit-per-pixel images in terms of units DHWT, Generally, when the PSNR is DHWT or greater, the original and rebuilt images are almost indiscernible to human eyes. The compression ratio, the original image, and the compressed image with DHWT, level 2 decomposition The results are shown in Table 2. As a result, in the preceding example, treating DHWT with the reference in rows and columns reveals that the approximate coefficients are concentrated in the first quarter, dubbed LL, and the detail coefficients are concentrated in the second, third, and fourth quarters, dubbed LH, HL, and HH, respectively. The detail coefficients are divided into three parts: in the LH band, they are called horizontal detail coefficients denoted by Dh, in the HL band, they are called vertical detail coefficients denoted by Dv, in the HH band, they are called diagonal detail coefficients denoted by Dd . After processing the original signal in the matrix I1 by using DHWT on the rows obtained I2 and on the columns obtained I3 it is divided into approximate coefficients and detailed coefficients; this operation is called de composition signal. The reconstructed signal is then obtained by using inverse DHWT. Figure 3 shows the process of analyzing the color image and the new wavelet effect.

**TABLE 2. Shows Effect of DHWT filter with loops of true compression b on Image**

| Loop | CINH | DCI | CI | MSE | PSNR | CR | BPP |
|---|---|---|---|---|---|---|---|
| 1 | 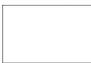 | 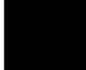 | 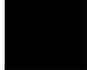 | $1.99e^{+04}$ | 5.142 | 0.03% | 0.007 |
| 2 | 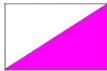 | 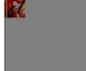 | 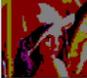 | 6744 | 9.841 | 13.52% | 3.245 |
| 3 | 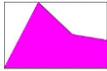 | 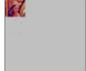 | 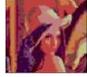 | 1530 | 16.28 | 14.58% | 3.524 |
| 4 | 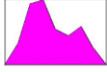 | 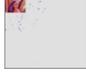 | 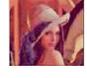 | 543.5 | 20.78 | 15.87% | 3.809 |
| 5 | 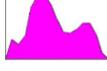 | 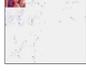 | 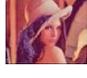 | 211.3 | 24.88 | 17.97% | 4.312 |
| 6 | 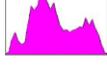 | 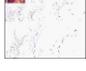 | 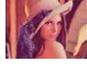 | 85.32 | 28.82 | 21.72% | 5.212 |
| 7 | 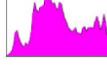 | 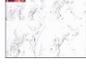 | 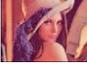 | 34.99 | 32.69 | 27.80% | 6.624 |
| 8 | 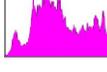 | 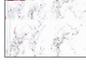 | 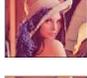 | 15.27 | 36.29 | 35.54% | 8.625 |
| 9 | 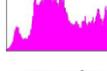 | 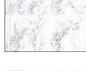 | 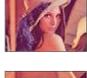 | 8.314 | 38.93 | 46.86% | 11.246 |
| 10 | 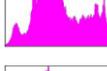 | 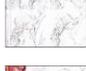 | 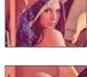 | 6.296 | 40.14 | 59.68% | 14.323 |
| 11 | 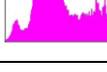 | 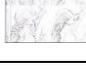 | 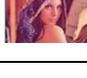 | 5.831 | 40.47 | 73.65% | 17.676 |

| 12 | 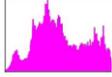 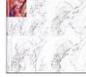 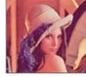 | 5.831 | 40.47 | 73.65% | 17.676 |

From level 1 to level 8, the proposed wavelets were tested to reach the best results from level 2, as shown in the table (3), which also shows a comparison of the results with a number of examples of the basic wavelets Symlet2, coiflet 2, and daubecheis2 for the same levels

TABLE 3. Compare DHWT with some basic wavelets

| | DHWT | | | |
|---|---|---|---|---|
| level | MSE | PSNR | BPP | CR |
| 1 | 8.058 | 39.072 | 25.102 | 46.26% |
| 2 | 5.831 | 40.47 | 17.676 | 73.65% |
| 3 | 4.58 | 41.33 | 11.102 | 46.26% |
| 4 | 3.505 | 41.59 | 7.344 | 30.60% |
| 5 | 3.49 | 42.55 | 8.650 | 33.38% |
| 6 | 2.77 | 43.25 | 9.718 | 33.65% |
| 7 | 1.94 | 44.33 | 13.406 | 52.86% |
| 8 | 1.93 | 44.33 | 13.406 | 55.0% |
| | SYM 2 | | | |
| level | MSE | PSNR | BPP | CR |
| 1 | 8.850 | 38.66 | 24.1 | 46.01% |
| 2 | 5.922 | 40.41 | 16.991 | 70.80% |
| 3 | 4.924 | 41.21 | 10.724 | 44.62% |
| 4 | 4.2 | 41.83 | 6.9 | 28.8% |
| 5 | 4.3 | 41.79 | 6.7 | 27.9% |
| 6 | 4.3 | 41.79 | 6.6 | 27.7% |
| 7 | 4.3 | 41.8 | 6.6 | 27.2% |
| 8 | 2.6 | 42.89 | 12.9 | 53.8% |
| | COIF 2 | | | |
| level | MSE | PSNR | BPP | CR |
| 1 | 9.8 | 38.1 | 24.0 | 100% |
| 2 | 6.468 | 40.02 | 16.320 | 68.00% |

| level | MSE | PSNR | BPP | CR |
|---|---|---|---|---|
| 3 | 8.5 | 38.8 | 7.5 | 31.44% |
| 4 | 3.9 | 42.`1 | 6.7 | 28.1% |
| 5 | 4.0 | 42.1 | 6.5 | 27.1% |
| 6 | 4.0 | 42.1 | 6.4 | 26.9% |
| 7 | 4.0 | 42.1 | 6.4 | 26.8% |
| 8 | 2.4 | 40.51 | 12.5 | 52.4% |

db2

| level | MSE | PSNR | BPP | CR |
|---|---|---|---|---|
| 1 | 10.0 | 38.1 | 22.5 | 93.8 |
| 2 | 5.992 | 40.41 | 16.991 | 70.8% |
| 3 | 9.5 | 38.3 | 8.1 | 34.0% |
| 4 | 4.5 | 41.5 | 7.3 | 30.6% |
| 5 | 4.5 | 41.5 | 7.1 | 29.7% |
| 6 | 4.8 | 41.2 | 7.0 | 29.4% |
| 7 | 4.8 | 41.2 | 7.0 | 29.4% |
| 8 | 2.7 | 4.3.7 | 9.9 | 41.4% |

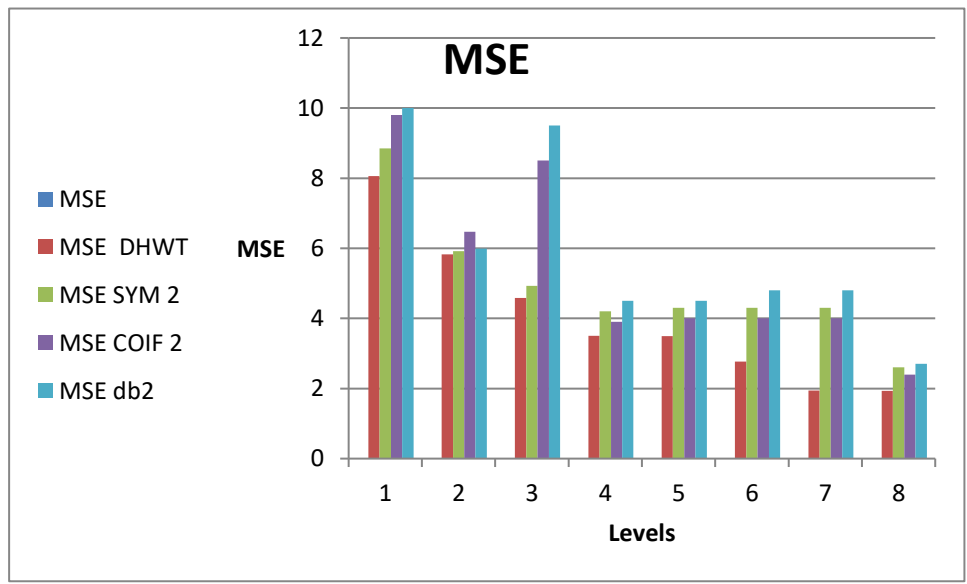

FIGURE5: Comparison MSE between the proposed wavelet and other wavelets

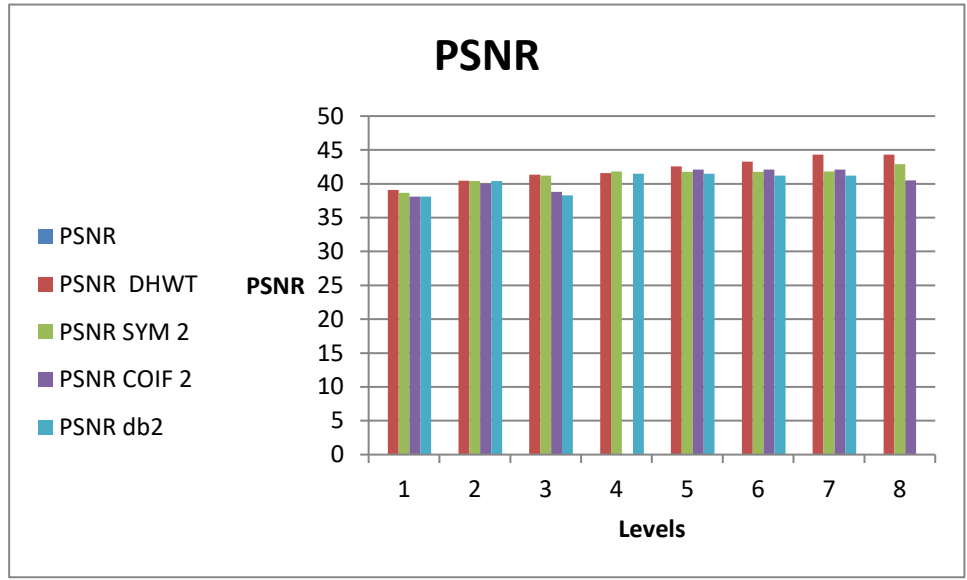

FIGURE 6: Comparison PSNR between the proposed wavelet and other wavelets

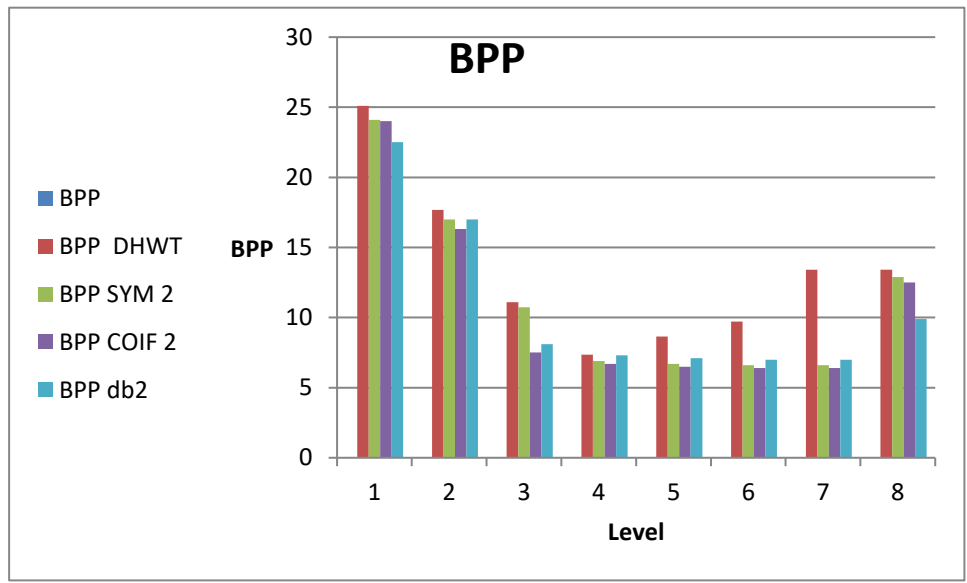

FIGURE 7: Comparison BPP between the proposed wavelet and other wavelets

**CONCLUSION**

Transferring information and sending pictures requires a large area that causes problems in transferring information, which led to the need to find rapid techniques to reduce this space and transfer information without loss. One of these techniques is image compression. In this work, Hermit constructed new wavelengths from orthogonal polynomials and proved their efficiency in the field of image processing. The lowest error is obtained from table (2). The image information is obtained in the eleventh step, where the same results are obtained in the twelfth and thirteenth steps to the nineteenth step, compared to previous results with Symlet 2, Conflict 2, and Daubecheis 2, and duplicate results are obtained in the sixteenth step. Table (3) shows the efficiency of the proposed wavelets, where the best results were obtained at the second level compared to what the basic wavelets Symlet 2, Conflict 2, and daubecheis2 reached at the same level, which took more time to reach good results.